\documentclass[letterpaper, 10pt, conference]{ieeeconf}
\IEEEoverridecommandlockouts
\let\labelindent\relax
\usepackage{enumitem}

\usepackage{booktabs}
\usepackage{cite}
\usepackage{csquotes}
\usepackage{amsmath,amssymb,amsfonts}
\usepackage{algorithmic}
\usepackage{graphicx}
\usepackage{textcomp}
\usepackage[table]{xcolor}
\usepackage{verbatim}
\usepackage[hidelinks]{hyperref}
\usepackage{fancybox}

\usepackage{tabularx}
\newcommand{\presponse}[2]{\enquote{#2} -- P#1}
\newcommand{\presponseblock}[2]{\begin{quote}\enquote{#2} -- P#1\end{quote}}

\newcommand{\themerow}[5]{#2 & \emph{#1} & #3 & \presponse{#4}{#5}\\}
\newcommand{\themeheader}{
  \rowcolors{2}{gray!25}{white}
  \begin{tabularx}{\linewidth}{>{\centering\arraybackslash}p{0.01\textwidth}p{0.20\textwidth}|p{0.3\textwidth}|X}
  \rowcolor{lightgray!50}
  \toprule
  & \textbf{Challenge} & \textbf{Description} & \textbf{Representative quote} \\
  \midrule}
\newcommand{\themefooter}{
  \bottomrule
  \end{tabularx}}

\newcommand{\insight}[1]{
    \boxedText{\textbf{Key Insight:} #1}}

\newcommand{\boxedText}[1]{
  \setlength{\fboxsep}{0.8em}
  \vspace{0.5em}
  \begin{center}
  \cornersize{.2}
  \Ovalbox{\begin{minipage}{0.9\linewidth}
    #1
    \end{minipage}}
  \end{center}
  \vspace{0.5em}}

\newcommand{\afs}[1]{{\color{orange}\textbf{(AA:} #1\textbf{)}}}
\newcommand{\ct}[1]{{\color{blue}\textbf{(CT:} #1\textbf{)}}}

\newcommand{\todo}[1]{{\color{red}(TODO: #1)}}

\title{A Study on the Challenges of Using Robotics Simulators for Testing}

\author{Afsoon Afzal$^{1}$
  and Deborah S. Katz$^{1}$
  and Claire Le Goues$^{1}$
  and Christopher S. Timperley$^{1}$
\thanks{$^{1}$School of Computer Science,
        Carnegie Mellon University, Pittsburgh, PA, USA
        {\tt\small afsoona@cs.cmu.edu, dskatz@cs.cmu.edu clegoues@cs.cmu.edu, ctimperley@cmu.edu}}%
\thanks{The first two authors contributed equally to this work.}
}

\begin{document}
\maketitle

\begin{abstract}
Robotics simulation plays an important role in the design, development, and
verification and validation of robotic systems.
Recent studies have shown that simulation may be used as a cheaper,
safer, and more reliable alternative to manual, and widely used, process of field testing.
This is particularly important in the context of continuous integration
pipelines, where integrated automated testing is key to reducing costs while
maintaining system safety. 
However, simulation and automated testing are not seeing the degree of widespread
adoption in practice that their potential would motivate. 
Our goal in this paper is to develop a principled understanding of
the ways developers use simulation in their process, and the challenges they
face in doing so.  This type of understanding can guide the development of more
effective simulators and testing techniques for modern robotics development. 

To that end, we conduct a survey of 82 robotics developers from a diversity of
backgrounds that addresses the current capabilities and limits of simulation
technology in practice.  
We find that simulation is used by 85\% of our participants for testing,
and that many participants desire to use simulation as part of their
test automation.
We identify 10 high-level
challenges that impede developers from using simulation
for manual and automated testing, and general purposes.
These challenges include the gap between simulation and reality,
a lack of reproducibility,
and considerable resource costs
associated with using simulators.
Finally, we outline avenues for improvement in the development of new simulators
that can help simulation reach its potential as a means of
verification and validation.
\end{abstract}

\section{Introduction}
\label{sec:intro}

Robotics simulators are invaluable tools that allow developers to rapidly
and inexpensively
design, prototype, and test robots in a controlled environment
without the need for physical hardware.
Popular simulators, such as Gazebo~\cite{gazebo}, V-REP~\cite{vrep},
and Webots~\cite{webots}, have been used to simulate a variety of systems
including industrial robots, unmanned aerial vehicles, and autonomous
(self-driving) cars.


Simulation is particularly promising for verification and
validation (V\&V) of robotic systems, 
potentially providing an automated, cost-effective,
and scalable alternative to the manual and expensive process of field
testing~\cite{NavigationBugs,TimperleyArdu2018,Robert2020,Gladisch2019}.
Simulation can effectively and automatically 
discover bugs in a variety of robot
application domains~\cite{asfault,Mullins2017,Tuncali2016,Rocklage2017}.
Numerous companies involved in the autonomy sector,
such as Uber~\cite{UberSimulation},
NVIDIA~\cite{NVIDIAConstellation},
and Waymo~\cite{WaymoReport},
use simulation on a large scale to develop, train, and test their
algorithms. 
The high demand for simulation in this sector has led to the development of 
a new generation of
specialized simulators,
such as CARLA~\cite{CARLA}, LGSVL~\cite{LGSVL}, AirSim~\cite{airsim}, and AADS~\cite{AADS}.


However, simulation-based testing does not dominate robotics V\&V to a degree
commensurate with its potential.  Instead, 
field testing remains the predominant means of V\&V
for robotic systems~\cite{ZhengState2017,AfzalICST20}.
Studies have compared robotics simulators on aspects such as their
features, usability, performance, documentation, and graphical user 
interface (GUI)~\cite{simulator-comparison,simulator-features-compared}.
However, these studies do not answer the question of why simulation is not more
widely adopted as a core V\&V practice, or what challenges developers face when
using it.  Indeed, prior work studying
the challenges of
testing in robotics~\cite{AfzalICST20} and cyberphysical systems (CPSs)~\cite{ZhengState2017}
in general identify simulation as a key element of robotics/CPS testing that
requires improvement. 

Our goal in this paper is to develop a grounded understanding of
the ways developers use simulation in their process and the challenges they
face in doing so. 
This type of understanding can guide the development of more
effective simulators and testing techniques for modern robotics development
that are better suited to developer needs and that can ultimately result in
higher quality robots.

To this end, we conduct a study of robotics developers to understand
how they perceive simulation-based testing, and what 
challenges they face while using simulators.
Our survey with 82 participants confirms that simulation is a popular
tool among robotics developers and that testing is its most common use case.
From our participants' responses, we identified 10 challenges
that make it difficult for developers to use simulation in general, for testing, and
specifically for automated testing.
The general challenges
include the difficulties of learning and using simulators,
the lack of realism, and the absence of specific capabilities, which constrain
the way developers use simulation.
The challenges 
that limit the extent of simulation-based testing
include a lack of reproducibility, the complexities
of scenario and environment construction, and considerable resource costs.
Finally, the absence of automation features, a lack of reliability, and
API instability discourage developers from using simulation for test automation
and prevent developers from realizing the benefits of continuous integration.
We believe that the results of this study can inform the construction
of a new generation of software-based simulators, designed to better accommodate
the needs of developers that arise during robotics testing.



Overall, we make the following contributions:
\begin{itemize}
\item We conduct a study of 82 robotics developers from
a variety of organizations and with diverse levels of experience in 
robotics.
\item We find that developers are using simulation extensively for testing
  their robots and that many developers want to incorporate simulation into their
  test automation.
\item We identify and explore ten key challenges that impede or prevent
  developers from using simulation in general and for manual and
  automated testing.
\item We suggest opportunities for improvement
  that may address the identified challenges.
\item We provide our survey materials and additional results to allow the community
  to build on our research.
\end{itemize}

\section{Methodology}
\label{sec:methodology}

In this study, we aim to better understand the ways in which robotics developers use
simulation as part of their testing process, and the challenges they face in doing so,
by addressing the following research questions:

\begin{description}
\item[\textbf{RQ1:}] What challenges do developers face when using simulation
  in general?
\item[\textbf{RQ2:}] What challenges do developers face when using simulation
  for testing?
\item[\textbf{RQ3:}] What challenges do developers face when using simulation
  for test automation?
\end{description}

To answer these 
questions, we conducted an open-ended online survey
of robotics developers in November 2019.




To reach our intended audience (i.e., robotics developers), we distributed our survey
via social media outlets, email, and several popular forums within the community:
We posted to the ROS 
and Robotics
subreddits on Reddit,\footnote{\url{https://reddit.com}}
the ROS Discourse,\footnote{\url{https://discourse.ros.org}}
and the RoboCup forums.\footnote{\url{http://lists.robocup.org/cgi-bin/mailman/listinfo}}
We also advertised our survey on Facebook and Twitter,
and posted a recruitment email to the Carnegie Mellon Robotics Institute and
National Robotics Engineering Center mailing lists.

In total, 151 participants took the survey, out of which 82 completed it.
For the purpose of analysis, we only consider the 82 completed responses.
All 82 participants that completed the survey reported that they had used
a robotics simulator.
%
Figure~\ref{fig:demographics:experience} provides an overview of the demographics of
the 82 participants that completed the survey.
In terms of experience, more than two thirds of participants (71.95\%) reported
having worked with robotics software for more than three years.
Most participants (79.27\%) reported that they had worked with robotics in academia at some point
during their life, and almost two thirds (65.85\%) reported working with robotics in industry
at some point. Participants report that they currently work at organizations of
varying sizes.
Overall, our study sample is composed of a diverse array of candidates with differing
levels of experience that have worked in a variety of organizations, thus
ensuring that the results of the study are not limited to any one population.

\begin{figure*}[t]
\centering
\begin{tabular}{lrr lrr lrr}
\toprule
\textbf{Experience} & & & \textbf{Organization} & & & \textbf{Size of organization} & & \\
Years of experience & \# & \% & Type & \# & \% & Number of people & \# & \% \\
\midrule
Less than one year          & 10 & 12.20\%  & Academia    & 65 & 79.27\%  & 1--10 people          & 22  & 26.83\% \\
Between one and three years & 13 & 15.85\%  & Industry    & 54 & 65.85\%  & 11--50 people         & 23  & 28.05\% \\
Between three and ten years & 40 & 48.78\%  & Individual  & 35 & 42.68\%  & 51--100 people        & 9   & 10.98\% \\
More than ten years         & 19 & 23.17\%  & Government  & 12 & 14.63\%  & More than 100 people  & 28  & 34.15\% \\
                            &    &          & Other       & 9  & 10.98\%  &                       & \\
\bottomrule
\end{tabular}
\caption{Demographics for the 82 survey participants that completed the survey
  in terms of their experience, the types of organization at which they had worked,
  and the size of the most recent organization to which they belonged.}
\label{fig:demographics:experience}
\end{figure*}



To analyze the open-ended survey responses, we used descriptive
coding~\cite{Saldana15} to assign one or more short labels to each segment of
data, identifying the topic(s) of that segment.
After developing an initial set of codes, we conducted a process of
adjudication to reach consistency and agreement, before using code mapping to
organize the codes into larger categories~\cite{Saldana15,team-codebooks,Charmaz14}.
Finally, we used axial coding to examine relationships between
categories and to identify a small number of
overarching research themes.

To facilitate data reuse and to aid others in the reproduction of our results,
we share the recruitment materials, questionnaire, and codebook for our
study at the following URL:
\url{https://bit.ly/2wRuEFP}.\footnote{We provide access to
anonymized survey responses upon request.}

%


\section{Results}
\label{sec:results}

\begin{figure}[t]
\centering
\includegraphics[width=0.5\textwidth]{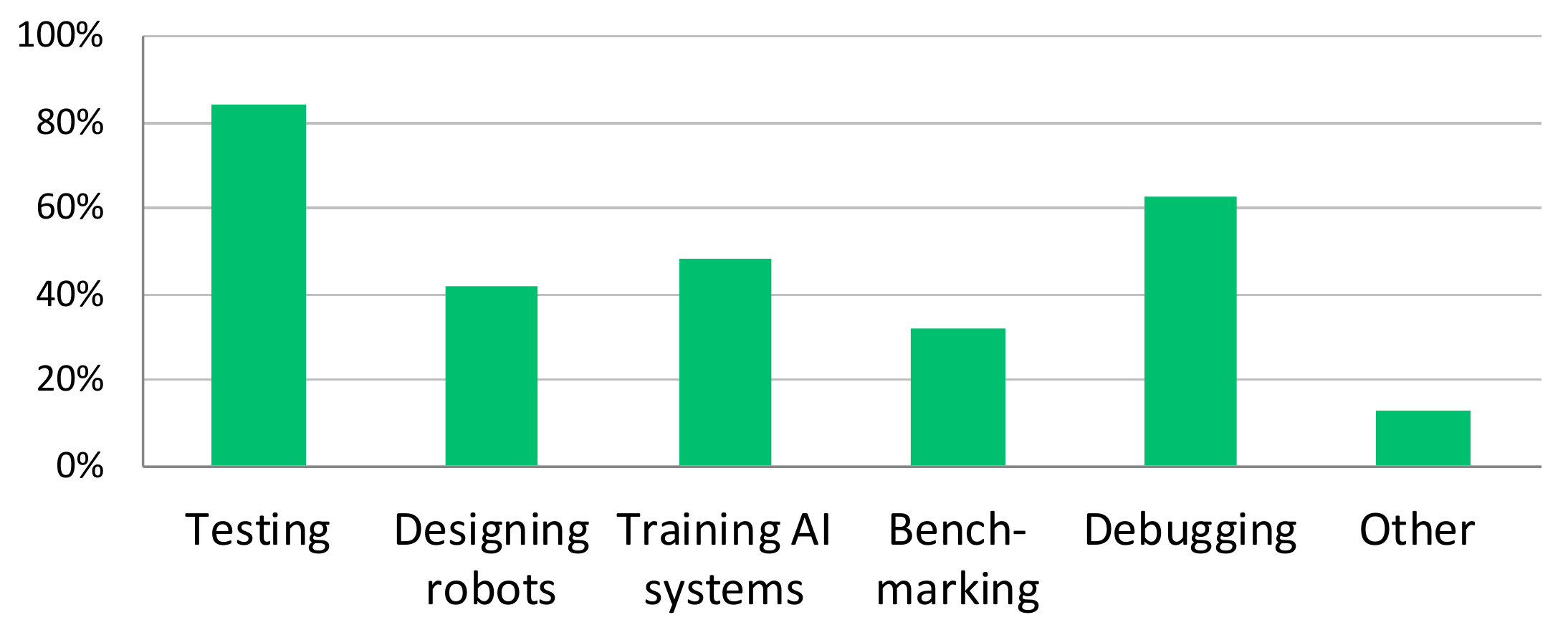}
\caption{An overview of the high-level reasons that participants gave for
  using simulation (82 responses).}
\label{fig:simulator-purposes}
\end{figure}

\newcommand{\general}{$\blacksquare$}
\newcommand{\testing}{$\blacktriangle$}
\newcommand{\automation}{$\bigstar$}

\begin{figure*}[t]
\centering
\themeheader
  \themerow{Reality gap}
  {\general}
  {The simulator does not sufficiently replicate the real-world behavior of the robot to a degree that is useful.}
  {33}
  {[Simulation is n]ot realistic enough for accurately modeling task; preferred running on real robot}

  \themerow{Complexity}
  {\general}
  {The time and resources required to setup a sufficiently accurate, useful simulator could be better spent on other activities.}
  {4}
  {It was easier and more accurate to setup and test on a physical system than simulate}

 \themerow{Lacking capabilities}
  {\general}
  {Simulators may not possess all of the capabilities that users desire, or
  those simulators that do may be prohibitively expensive.}
  {77}
  {[M]ost simulators are good at one thing, some are good at simulating the vehicles (drone,robot,car,etc) some are good at simulating the environment (good for generating synthetic data) some are good at senors, some are good at physics, some are good at pid control, etc. but not one has all these attributes.}



  \themerow{Reproducibility}
  {\testing}
  {Simulations are non-deterministic, making it difficult to repeat simulations, recreate issues encountered in simulation or on real hardware, and track down problems.}
  {42}
  {Deterministic execution: the same starting conditions must produce absolutely identical results.}


 \themerow{Scenario and environment construction}
  {\testing}
  {It is difficult to create the scenarios and environments required for testing the system in simulation.}
  {38}
  {Setting up a simulation environment is too much work, so I don't do it often.}

  \themerow{Resource costs}
  {\testing}
  {The computational overhead of simulation requires special hardware and computing resources which adds to the financial cost of testing.}
  {37}
  {Simulating multiple cameras (vision sensors) with full resolution at a high frame rate is usually very slow and therefore not practical.}

\themerow{Automation features}
  {\automation}
  {The simulator is not designed to be used for automated testing and does not allow headless, scripted or parallel execution.}
  {34}
  {Most simulations are NOT designed to run headless, nor are they easily scriptable for automatic invocation.}

 \themerow{Continuous integration}
  {\automation}
  {It is difficult to deploy the simulator in suitable environments for continuous integration (e.g., cloud computing servers).}
  {62}
  {The simulation requires some computational resources which can be difficult to be part of CI, especially when our CI is running on the cloud}

  \themerow{Simulator reliability}
  {\automation}
  {The simulation is not reliable enough to be used in test automation in terms of the stability of the simulator software, and the timing and synchronization issues introduced by the simulator.
}
  {80}
  {There were many challenges -   1. Getting difference in the real time and simulation time   2. Changing the entire physics engine source code for our application  3. Glitch during the process of trying to move the real hardware with the simulation model.}

\themerow{Interface stability}
  {\automation}
  {The simulator's interface is not stable enough or sufficiently well-documented to work with existing code or testing pipelines.
}
  {28}
  {[We have automation difficulties with] integration into existing code, missing APIs, stability of libraries}

\bottomrule
\end{tabularx}
\caption{Summary of challenges participants encountered
  when using simulation in general (\general), specifically for testing (\testing), and for test automation (\automation).}
\label{fig:challenges}
\end{figure*}


Our survey asked participants broadly about their use of simulation.
We find that our participants are unanimously familiar with simulation, and
they use it on a
regular basis for a variety of important purposes:
59 out of 82 (71.95\%) participants reported that they used simulation within the
last month at the time of completing the survey.
When asked about their most recent project that involved simulation, 51 of
82 (62.20\%) participants reported that they used a simulator daily, and
25 of 82 (30.49\%) participants reported that they used a simulator on a weekly basis.

Figure~\ref{fig:simulator-purposes} presents the variety and popularity of
purposes for which our participants use simulation.
Almost 85\% of participants have used simulation for testing, and
testing is the most popular use case for simulation.
When asked for details on how they use simulation for testing, participants 
reported using it for testing algorithms, variability
testing, component testing, sanity checking, and multi-robot testing.
This finding suggests that developers generally see value in
using simulation for testing.


Participants also reported that they use simulation for testing
when it is unsuitable or impractical to 
test on real hardware or in a real environment.
They reported using simulation to better understand the design and
behavior of existing robotic systems and their associated software, and
to incorporate simulation into automated
robotics testing, including continuous integration (CI).

Of the 85\% of participants who have used simulation for testing, we find that
roughly 60\% of them also have tried to use simulation as part of their test automation.
These findings demonstrate that developers find simulation to be a valuable tool
for testing, and there is a desire to incorporate simulation-based testing into
their test automation processes.

Given the ubiquity of simulation and its importance to robotics testing and
development, it is vital that we, as a community, understand the barriers that
developers face when using simulation.
By bringing these barriers to the attention of the community, we can work to
lower those barriers and empower developers, bringing us closer to the potential
of simulation, and advancing the state of
robotics software development and quality assurance.


\insight{Simulation is an essential tool for developers that is used extensively for
  building and testing robot software.
  Given its importance, it is vital that we better understand
  the challenges that prevent developers from realizing its full potential.}

In the following sections, we present the challenges of using simulators, given
in Figure~\ref{fig:challenges}.
Section~\ref{sec:results:rq1} discusses challenges that apply broadly to many uses of simulation,
Section~\ref{sec:results:rq2} narrows the focus to those challenges that apply when simulation
is used for testing,
and Section~\ref{sec:results:rq3} further narrows to the challenges specific to test automation.

\subsection{RQ1: What challenges do developers face when using simulation in general?}
\label{sec:results:rq1}

\begin{figure}[t]
\centering
\begin{tabular}{p{0.65\columnwidth}rr}
\toprule
\textbf{Reason for not using simulation} & \textbf{\#} & \textbf{\%} \\
\midrule
Lack of time or resources & 15 & 53.57\% \\
Not realistic/accurate enough & 15 & 53.57\% \\
Lack of expertise or knowledge on how to use software-based
simulation
& 6 & 21.43\% \\
There was no simulator for the robot & 4 & 14.29\% \\
Not applicable & 4 & 14.29\% \\
Too much time or compute resources & 2 & 7.14\% \\
Nobody suggested it & 0 & 0.00\% \\
Other & 2 & 7.14\% \\
\bottomrule
\end{tabular}
\caption{An overview of the reasons that participants gave for not using simulation
  for a particular project, based on 28 responses.}
\label{fig:reasons-for-not-using-simulation}
\end{figure}

Although we find that simulation is popular among developers, 28 of 82 (34.15\%)
participants reported making a decision to not use simulation for
a project for a variety of reasons, given in Figure~\ref{fig:reasons-for-not-using-simulation}.
By analyzing both these reasons
and the difficulties that participants experienced when they did use simulation,
we identified three high-level challenges of using simulation
in general, discussed below.

\begin{description}[leftmargin=1.2em]
\item[\textbf{Reality gap:}]
  A number of participants cited an inadequate representation of physical reality (i.e., the reality gap)
  as both a challenge when trying to
  use simulation, and a reason not to use it in the first place.
  P33 notes that simulation can produce unrealistic behaviors that would
  not occur in the real world.
  P16 highlighted that accounting for all relevant physical phenomena can also
  be challenging:
  \enquote{my simple simulation model did not include a tire model,
    so simulations at higher speeds did not account for realistic behaviors
    for cornering or higher accelerations or deceleration.}
  In particular, realistically modeling stochastic processes (e.g., signal noise)
  and integrating those models into the simulation as a whole is a challenge:
  P15 shared,
  \enquote{A classic problem is integrating wireless network
    simulation with physical terrain simulation. This also applies to GPS
    signal simulation, as well.}

  For some, such as P29, the reality gap can be too large to make simulation
  valuable: \enquote{too big discrepancy between simulation results and reality
  (physical interaction).}
  For others, if not many, simulation can still serve as a valuable tool
  despite the existence of the reality gap. As P36 puts it,
  \enquote{Software behavior in sim is different compared to real, so not
    everything can be tested, but a lot can be.}


%
\item[\textbf{Complexity:}]
Accurate simulation of the physical world is an inherently challenging process that
naturally involves a composition of various models.
Alongside the essential complexity of simulation are sources of
\emph{accidental complexity}~\cite{nosilverbullet} that do not relate to the
fundamental challenges of simulation itself, but rather the engineering
difficulties that developers face when trying to use simulation.
These sources of accidental complexity may ultimately lead users
to abandon or not use simulation at all.
Inaccurate, inadequate, or missing documentation can make it difficult to learn
and use a simulator, as P22 highlighted:
\enquote{Lack of documents for different platform types and sometimes wrong
documentation makes us lose a lot of time working on [stuff] that will never work,
for example, the Gazebo simulator does not work well in Windows.}
In some cases, documentation may be written in another language;
P74 stopped using simulation for a project for that reason:
\enquote{The language was Japanes[e], but we don't speak that language so we
  couldn't use well the simulator.}

Difficult-to-use APIs make it difficult to extend the simulator with new
plugins, and a lack of support for industry-standard 3D modeling formats
in widely used simulators such as Gazebo makes creating models a tedious
and error-fraught process:

\presponseblock{4}{Gazebo is the de-facto [simulator] right now and is poorly
  documented and difficult to customize to any degree.}

Together, these sources of complexity increase the learning curve of many
simulators and may lead developers to abandon or not use them in the first
place. P20 said, \enquote{Steep learning curve in understanding the test
environment software setup and libraries. Without a good software engineering
skills the simulated environment will not replicate the real environment.}


\item[\textbf{Lacking capabilities:}]
Finding a simulator that provides all of the characteristics a user desires
can be challenging. P77 highlighted that while it is possible to
find a simulator that is good in one particular aspect, it is hard to find a
simulator that is good in all desired aspects.

As P4 pointed out, simulators that do possess all of the desired qualities
also tend to be very expensive:
\enquote{Adding plugins is usually very challenging, and the only good
  frameworks that do any of this stuff well are very expensive (V-Rep and
  Mujoco for example).}

We asked which simulation features participants desired most but are unable
to use in their current setups.
Among the most important features they mentioned were
the ability to simulate at faster-than-real-time speeds,
native support for headless execution (discussed in Section~\ref{sec:results:rq3}),
and an easier means of constructing environments and scenarios
(discussed in Section~\ref{sec:results:rq2}).

Numerous participants desired the ability to run simulation at faster-than-real-time
speeds but were unable to do so in their current simulation setups.
For example, P52 said, \enquote{W[e] needed to speed up simulation time, but that was
difficult to achieve without breaking the stability of the physics engine.}
The ability to run at faster-than-real-time speeds is useful not only for
reducing the wall-clock time taken to perform testing, but for other purposes,
as P62 highlighted:
\enquote{Faster than real time is really important to produce training data for deep learning.}

Several participants also desired particular features that would increase the overall fidelity of the simulation. P46 wanted support for
\enquote{Advanced materials in environments (custom fluids, deformable containers, etc.).}
  Interestingly, P69 desired the ability to tune the fidelity of the simulation:
  \enquote{Ability for controllable physics fidelity. First order to prove
  concepts then higher fidelity for validation. Gazebo doesn’t have that.}
  Recent studies have shown that low-fidelity simulation can be used as an effective
  and inexpensive way of discovering many bugs in a resource-limited
  environment~\cite{NavigationBugs,TimperleyArdu2018,Robert2020}.


Other capabilities specified by participants include native support for
multi-robot simulation and
large environments,
and the ability to efficiently distribute a simulation session across multiple
machines.

%
\end{description}


Ultimately, the complexities of setting up and using simulation,
the reality gap, and the time and resources necessary
to make the simulation useful led some participants to use physical
hardware instead. As P4 said,
\enquote{It was easier and more accurate to setup and test on a physical system
  than simulate.}

\insight{
  Developers find considerable value in simulation, but
  difficulties of learning and using simulators,
  combined with a lack of realism and specific capabilities,
  constrain the way that developers use simulation.
  By alleviating these challenges, simulation can be used
  for a wider set of domains and applications.

}

\subsection{RQ2: What challenges do developers face when using simulation for testing?}
\label{sec:results:rq2}

Participants reported a variety of challenges in attempts to use simulation for testing,
summarized in Figure~\ref{fig:challenges}.
We identified the following challenges that mainly affect the use of
simulation for testing:
\begin{description}[leftmargin=1.2em]
\item[\textbf{Reproducibility:}]
The lack of reproducibility and presence of non-determinism in simulators 
lead to difficulties when testing, as reported by participants.
P42 highlighted
that a \enquote{Lack of
deterministic execution of simulators leads to unrepeatable results.}
This points to a need to accurately reproduce system failures that are
discovered in testing, in order
to diagnose and debug those failures.
If a tester cannot consistently reproduce the failures detected in
simulation, it will be difficult to know whether changes made to the code
have fixed the problems.
P7 pointed to the particular difficulty with achieving reproducibility in Gazebo: \enquote{Resetting gazebo simulations was not repeatable enough to get good data.} 
P48 and P81 also mentioned a desire for reproducibility.

Consistent and systematic testing procedures rely on deterministic test outcomes.
This is particularly the case when incorporating test automation and continuous integration tests, which rely on automatically detecting when a
test has failed, as a sign that there is a problem with software changes.
Flaky~\cite{MiccoFlaky2016}
and non-deterministic tests may lead to the false conclusion that
a problematic software change does not have a problem (a false negative)
or that a good change has problems (a false positive).

\item[\textbf{Scenario and environment construction:}]
Testing in simulation requires a simulated environment and a test scenario.
Participants reported difficulty in constructing both test scenarios and
environments.
P38 said: \enquote{Setting up a simulation environment is 
too much work, so I don't do it often,} and P3 contributed, \enquote{Scripting scenarios was not easy. Adding different robot dynamics was also not easy.}
They wanted to be able to construct these more easily or automatically.
Participants pointed out that the scenarios or environments they require
sometimes must be created \enquote{by hand,} which requires a heavy time
investment and is subject to inaccuracies.
P4 said, \enquote{Making URDF files is a tremendous pain as the only good way to do it right now is by hand which is faulty and error prone,}
while P67 wanted, \enquote{Automated generation of simulation environments under some [custom] defined standards,} because
\enquote{The automated simulation environment generation is not easy. Plenty of handy work must be done by human operators.}


\item[\textbf{Resource costs:}]
Simulation is computationally intensive. It often benefits from specialized hardware, such as GPUs.
Participants report that these hardware requirements contribute strongly
to the expense of simulation.
These costs are compounded when tests are run multiple times, such as in
test automation.
For example, P42 reported that difficulties in using simulation as a part of
test automation include: \enquote{High hardware requirements (especially
GPU-accelerated simulators) driving high cloud server costs.}
Participants reported difficulties with running simulations in parallel or
taking advantage of distributed computing across several machines.
Participants also reported challenges in simulating large environments and
simulations of long duration, as they became too resource demanding to 
be practical. P67 requested,
\enquote{High computational performance when the environment size grows large (Gazebo performance drops down rapidly when the number of models raises).}
Participants also had issues relating to the cost of obtaining licenses
for appropriate simulators.
P66 reported that cost drove the choice not to use a particular simulator:
\enquote{Back then, Webots was not free,}
and P1 complained: \enquote{Not to mention the licensing price for small companies.}
\end{description}

\insight{
  Almost 85\% of participants used simulation for testing, but
  a lack of reproducibility,
  the complexities of scenario and environment construction,
  and considerable resource costs
  limit the extent of such testing.
}

\subsection{RQ3: What challenges do developers face when using simulation for test automation?}
\label{sec:results:rq3}

Research has shown that test automation can provide many benefits, including
cost savings and higher software quality~\cite{automated-testing-benefits}.
Despite the benefits of test automation, 27 of 69 (39.13\%) participants
reported never attempting to use simulation for this purpose.
Responses indicated that the challenges with using simulation, both in general
and for testing, prevented participants from attempting to incorporate it into
test automation.
Their reasons fell into three general categories:
\begin{enumerate}
\item Lack of value, where they
did not find test automation valuable or necessary. As P24 mentioned
\enquote{There were no obvious test harnesses available for the simulation 
environments I use and it did not seem obviously valuable enough to 
implement myself.}
\item Distrust of simulation, where they found the limitations of simulation to be too
restrictive to be used in test automation. 
\emph{Reality gap} and \emph{lacking capabilities} discussed in Section~\ref{sec:results:rq1} contribute to this belief.
P33 mentioned
\enquote{[Simulation is] not realistic enough for accurately modeling task; preferred running on real robot,} and P20 believed \enquote{Without a good software engineering skills the simulated environment will not replicate the real environment.}
\item Time and resource limitations, where the complexity of the simulator
(Section~\ref{sec:results:rq1}) and resource costs
(Section~\ref{sec:results:rq2}) prevented them from attempting test automation.
P14 explained \enquote{I didn't think of it most probably because I hadn't seen an example where software-based simulation was used for automated testing,}
and P17 simply reported \enquote{seemed very hard to do.}
\end{enumerate}

Among 42 people who have attempted using simulation as part of their test
automation, 33 (78.57\%) reported facing difficulties.
Based on their descriptions of these difficulties, we identified the
following four challenges specifically affecting test automation:

\begin{description}[leftmargin=1.2em]
\item[\textbf{Automation features:}]

Although a GUI is an important component of a simulator, participants 
reported a preference towards running the simulator headless (i.e., without the GUI)
when used for test automation.
Disabling the GUI eliminates the computational overhead of the simulator 
caused by rendering heavy graphical models.
Not being able to run the simulator headless
is one of the major difficulties our participants face for automation.
\presponseblock{37}{Making the simulator run without GUI on our Jenkins server\footnote{Jenkins is a continuous integration service: \url{https://jenkins.io}} turned out to be more difficult than expected. We ended up having to connect a physical display to the server machine in order to run the simulation properly.}

Furthermore, the ability to set up, monitor, and interact with
the simulation via scripting, without the need for manual intervention, is
vital for automation.
Our participants reported the need to devise creative solutions in the absence
of support for scripting.
P8 shared, \enquote{Ursim needs click-automation to run without
human interaction,} where they used an click-automation tool to be able to
run the simulator automatically. 

\item[\textbf{Continuous integration:}]

Continuous integration (CI) is emerging as one of the most successful techniques
in automated software maintenance.
CI systems are used to automate the building, testing, and deployment of software.
Research has shown that CI practices have a positive effect on software 
quality and productivity~\cite{ci-benefits}.

CI, by definition, is an automated method,
and in many cases involves the use of cloud services such as TravisCI.\footnote{\url{https://travis-ci.org}}
Our participants faced difficulties
engineering the simulation to be used in CI and run on cloud servers.
For example, P66 mentioned \enquote{I wasn't able to setup a CI pipeline which runs in GPU machines, for use with rendering sensors.}

Many of these difficulties arise from lacking automation features
(e.g., headless execution) and
high resource costs (e.g., requiring expensive, GPU-heavy hardware),
discussed earlier as challenges.
P77 reported
\enquote{It is expensive to spin up cloud GPU VMs to run the simulator.}


\item[\textbf{Simulator reliability:}]
One of the challenges of using a simulator in a test automation pipeline
is the reliability of the simulator itself. In other words,
participants reported facing unexpected crashes, and timing and
synchronization issues while using the simulator in automation.
P29, P54, P73, and P80 all reported software stability and timing issues
as difficulties they faced for automation.
P29 further elaborated difficulty in ensuring a clean termination of the simulator.
That is, when the simulator crashes, it should properly store the
logs and results before termination of the simulation,
and properly kill all processes to prevent resource leaks.
Clean termination is particularly relevant to test automation as
resource leaks may compound when simulations are repeated, up to the point
where it interferes with the ability to run additional simulations
and requires manual intervention.


\item[\textbf{Interface stability:}]

The stability of the simulator interface can have a significant impact
on the automation process because inconsistent simulator APIs can lead to failures
in client applications~\cite{api-stabiliity}.
Our participants reported unstable and fragile interfaces as a
challenge for automation. For example, P39 mentioned
\enquote{APIs are pretty fragile and a lot of engineering need to be done to get it working.}

Five participants reported difficulties in integrating existing code or
infrastructure with simulation APIs. P80 mentioned
changing the entire physics engine source code for their application.
More specifically, participants desired better integration of simulators 
with ROS. For example, P74 shared
\enquote{I would like that [the simulator] can be use with ROS.}

\end{description}

\insight{
Developers desire to include simulation as part of their test
automation, but most developers that attempt to do so
face numerous difficulties.
These difficulties include an absence of automation features, and a lack
of reliability and API stability.
Ultimately, these challenges discourage developers from using simulation
for test automation, limit the extent to which it is used,
and prevent
developers from leveraging the benefits of continuous integration.}

\section{Discussion}
\label{sec:discussion}



Despite the myriad benefits of using simulation for testing,
many popular simulators do not appear to make suitable accommodations
for that use case.
For example, participants report that Gazebo, the \emph{de facto} simulator for
the Robot Operating System~\cite{ros},
does not adequately support headless execution,
lacks reproducibility,
and performs poorly when used to simulate complex and large environments.
%


As robots and their associated codebases become larger and more complex,
the need for, and cost of, a continuous process of verification and validation
will increase considerably.
The popular but expensive practice of using field testing to assure correctness
will be unable to handle these increased needs by itself
due to practical limits on hardware, human resources,
and safety~\cite{AfzalICST20}.
Simulation-based testing
may serve as a cheaper, safer, and more reliable alternative
by addressing the challenges identified in this paper.

To achieve this potential:
(1) simulators should be made easier to use for both basic and advanced
purposes,
(2) simulators should ambitiously expand their capabilities to support 
complex, large-scale environments that better resemble the robots' deployment
in the physical world,
and (3) simulators should be built to support scalable automation.


Simulators could be made easier to use in general by eliminating sources of
complexity,
introducing user-friendly features, and improving documentation.
Examples of user-friendly features that participants requested
include: providing a web interface by default, rather than a traditional
graphical user interface; support for models written in industry-standard
formats; and augmented reality visualizations.
Such changes would allow more developers to reap the benefits of simulation
by reducing the learning curve, and would reduce the considerable
investment of time required to use simulation for more advanced
purposes such as automated testing.

The scope and capabilities of simulators could be expanded to
support larger, more realistic simulations of real-world robot
deployments for a wider range of domains.
To do so,
simulators must efficiently support
large, detailed environments that may contain multiple robots,
and achieve greater physical fidelity
without increasing resource costs.
Additionally, simulators should strive to provide powerful, interactive tools
that allow developers to easily design and generate vast
scenarios and environments.

To support scalable automation
of simulation-based testing as part of
continuous process of verification and validation, simulators should:
(a) provide support for headless execution, and scripting via stable
and well-designed APIs;
(b) ensure reproducible results and reliable simulation to allow developers to
quickly and easily investigate discovered failures;
and (c) substantially reduce the resource costs and hardware requirements of
simulation.
Addressing these needs would allow simulation to
be deployed inexpensively in cloud environments
as part of continuous integration.


\paragraph*{The state of the art}Significant progress towards these goals is being made by
a new generation of simulators. 
Several of these new simulators are specialized for particular domains:
CARLA~\cite{CARLA}, LGSVL~\cite{LGSVL}, and AADS~\cite{AADS} are specialized for automated driving
applications, and AirSim~\cite{airsim} simulates a wider variety of autonomous vehicles.
Notably, all of these simulators are built on top of popular video
game engines, and support complex, dynamic urban environments.
AADS~\cite{AADS} enhances the visual fidelity of simulation by
integrating photos, videos, and sensor readings, allowing for more realistic
testing of perception components.
In contrast to these specialized simulators, Ignition Gazebo~\cite{IgnitionGazebo},
the descendant of the Gazebo simulator, is agnostic to
application and domain. Instead, Ignition Gazebo's API supports
various rendering and physics backends, allowing the developer to customize
the simulator to better fit the needs of a particular use case (e.g., fidelity
and performance).
AWS RoboMaker~\cite{robomaker} 
is a web-based IDE, simulator, and fleet management front-end
designed to make it easier to develop, test, and
deploy robot applications.
RoboMaker internally builds on top of Gazebo by adding infrastructure
for parallel simulations and automatic hardware scaling, and providing
numerous prebuilt environments (e.g., indoor rooms, retail stores, and race tracks).
Although each of these simulators addresses at least one of our identified
challenges, they have yet to become widely adopted in the community, and
it is as yet unclear whether they address enough of developers' needs in the
right combinations to succeed.

\section{Conclusion}
\label{sec:conclusion}

In this paper, we conducted a study of 82 robotics developers
to explore how robotics simulators are used, and the challenges that developers
commonly face when using simulation for general purposes, 
testing, and test automation.
Our results indicate that simulation is a popular tool among robotics 
developers, and is commonly used for testing with 85\% of participants
reporting having used simulation for testing, 
60\% of whom have also used simulation as part of their test automation.
We identified 10 high-level challenges associated with the use of 
simulation, and discussed these challenges in detail.
We further outlined ideas on how the
community can tackle these challenges to unlock the full potential of
simulation-based testing.

\section*{Acknowledgment}

We would like to thank both the ROS and Reddit communities,
and in particular, Chris Volkoff and Olly Smith,
for their invaluable support in distributing our survey.

This research was partially funded by
AFRL 
and DARPA: 
the authors are grateful for their support. Any opinions,
findings, or recommendations expressed are those of
the authors and do not necessarily reflect those of the US
Government.

\bibliographystyle{IEEEtran}
\bibliography{robot_sim}

\end{document}